\pdfoutput=1

\documentclass[11pt]{article}

\usepackage[]{ACL2023}

\usepackage{times}
\usepackage{latexsym}
\usepackage{booktabs}
\usepackage[T1]{fontenc}

\usepackage[utf8]{inputenc}

\usepackage{microtype}
\usepackage{tcolorbox}
\usepackage{multirow}
\usepackage{multicol}
\usepackage{inconsolata}
\usepackage{graphicx}
\usepackage{tabularx}
\usepackage{paralist}
\usepackage{amsmath,xparse,mleftright}

\title{Injecting Salesperson's Dialogue Strategies in Large Language Models\\ with Chain-of-Thought Reasoning}

\author{Wen-Yu Chang\quad Yun-Nung Chen \\
  National Taiwan University, Taipei, Taiwan\\
  \texttt{r10946031@ntu.edu.tw}\quad \texttt{y.v.chen@ieee.org} \\}

\begin{document}
\maketitle
\begin{abstract}

Recent research in dialogue systems and corpora has focused on two main categories: task-oriented (TOD) and open-domain (chit-chat) dialogues. TOD systems help users accomplish specific tasks, while open-domain systems aim to create engaging conversations. However, in real-world scenarios, user intents are often revealed during interactions.
A recent study introduced SalesBot, which simulates dialogues transitioning from chit-chat to task-oriented scenarios to train sales agents. Unfortunately, the initial data lacked smooth transitions and coherent long-turn dialogues, resulting in poor naturalness in sales-customer interactions.
To address these issues, this paper presents SalesBot 2.0, an improved dataset. It leverages commonsense knowledge from large language models (LLMs) through strategic prompting.
Additionally, we introduce a novel model called \textsc{SalesAgent}, trained on salesperson's interactions, using chain-of-thought (CoT) reasoning. This model excels in transitioning topics, understanding user intents, and selecting appropriate strategies.
Experiments using diverse user simulations validate the effectiveness of our method in controlling dialogue strategies in LLMs. Furthermore, SalesBot 2.0 enhances coherence and reduces aggression, facilitating better model learning for sales-customer interactions.\footnote{\url{https://github.com/MiuLab/SalesAgent}}

\end{abstract}

\section{Introduction}
\label{intro}

Dialogue systems have undergone significant advancements due to improvements in modeling techniques and computing power. However, most research in this field has focused on two distinct areas: task-oriented dialogues (TOD) and open-domain dialogues, also known as chitchat systems. 
Benchmark large-scale datasets for TOD include SGD (schema-suided dialogue)~\cite{rastogi2020towards} and MultiWoz \cite{budzianowski2018large,zang2020multiwoz}, which contain annotated information on the user intents and dialogue states. In TOD, the agent's goal is to identify the user's intention and fulfill their task by the end of the dialogue.
Meanwhile, research on open-domain chitchat systems and datasets \cite{li-etal-2017-dailydialog,adiwardana2020humanlike,zhang2018personalizing, kim2022soda} aims to build models capable of engaging in free-form conversations. 
As pre-trained language models continue to improve, larger sets of dialogues are being used to train models with the ability to engage in free-form chatting \cite{zhang-etal-2020-dialogpt,roller-etal-2021-recipes}. 
Despite significant advancements in both areas, there has been a lack of integration between them, which is crucial for real-world applications.

Recently, efforts have been made to integrate TOD and open-domain dialogues. For instance, \citet{sun2020adding} incorporated chitchat responses into existing TOD datasets to enhance the conversation's naturalness and social engagement. Furthermore, there have been attempts to develop datasets and models capable of handling both TOD and chitchat scenarios. \citet{li2022enhancing} developed the PivotBot model, capable of handling three predefined scenarios. One scenario involves adding chit-chat dialogues as context to a TOD, while another includes chitchat dialogues to facilitate domain transition. In contrast, the third scenario involves incorporating chitchat content to enhance a TOD, similar to the approach taken in the ACCENTOR model \cite{sun2020adding}. 
However, these approaches assume that the user has an \emph{explicit} goal to accomplish before starting the conversation, and chitchat responses merely enrich the conversation. 
In real scenarios, user intents or preferences are \emph{gradually} revealed during the course of conversations~\cite{chiu2022salesbot,murakhovs2023salespeople}.
Therefore, our work focuses on identifying potential intents, which may be explicitly or implicitly expressed by the user, and smoothly pivoting the dialogue to the associated tasks.

With this idea,
\citet{chiu2022salesbot} first introduced a framework for generating data that transitions from chit-chat to TOD dialogues. They utilized two open-domain BlenderBots to chat with each other and generate chit-chat dialogues, followed by a task-oriented intent detector to determine whether a transition to the TOD system should be made.
However, the SalesBot dataset has some limitations, such as the absence of proper social engagement in 20\% of the data, nonsensical detected intents due to short chit-chat context, and unnatural transition turns.
Our paper aims to improve the dataset by leveraging commonsense knowledge encoded in large language models (LLMs) to generate more human-like chit-chat dialogues, along with smooth intent-guided transition turns. 
Moreover, we propose a novel approach, \textsc{SalesAgent}, to inject dialogue strategies in LLMs by fine-tuning understanding and policy results from the improved data, SalesBot 2.0.
By incorporating chain-of-thought reasoning~\cite{wei2023chainofthought} for training LLMs, our model can follow the desired dialogue strategies and provide controllable and explainable interactions. 
The experiments demonstrate that our model trained on SalesBot 2.0 demonstrates superior interaction performance compared to a model trained on SalesBot 1.0 across all evaluated perspectives.

Our contributions can be summarized below:
\begin{compactitem}
\item This paper introduces an improved dataset encompassing open-domain chit-chats, seamless transitions, and potential intent detection from contextual cues.
The quality of the proposed dataset is demonstrated significantly better than the prior version in diverse perspectives.
\item This work is the first work that injects dialogue strategies in an end-to-end LLM for better controllability and explainability, where it is equipped with the integrated capabilities of intent detection, policy selection, and response generation.
\item This paper conducts a comprehensive evaluation of the proposed \textsc{SalesAgent}, comparing its performance with counterparts trained on the previous dataset across various perspectives at both turn-level and dialogue-level.
\end{compactitem}

\section{Related Work}
Our study focuses on a conversation scenario where a conversational agent attempts to steer the discussion towards determining whether the user is interested in receiving recommendations. Similar scenarios have been explored in various domains.

\paragraph{Persuasive Dialogue Dataset}
Prior studies \cite{hiraoka-etal-2014-reinforcement, yoshino-etal-2018-dialogue, wang-etal-2019-persuasion} focused on persuasive dialogue construction in different scenarios. \citet{hiraoka-etal-2014-reinforcement} annotated 34 dialogues in which a salesperson with expertise attempted to persuade a customer to purchase a camera. \citet{yoshino-etal-2018-dialogue} generated 200 dialogues through crowdsourcing, where one participant persuades the other to adopt a suggestion, such as cleaning a room. In comparison, \citet{wang-etal-2019-persuasion} collected 1,017 dialogues, where one participant was convinced to donate to a specific charity.

While all of these datasets are limited to specific scenarios, our framework can generate dialogues with any potential intent, making it more versatile. Additionally, our dataset is much larger, two to three times bigger than the previous ones, which makes it a valuable resource for training and evaluating non-collaborative conversational agents.

\paragraph{Conversational Recommendation}
Previous studies have developed various datasets for conversational recommendation systems. For instance, \citet{li2019deep} created a large-scale dataset with a focus on recommendation. Other researchers have utilized knowledge graphs to collect dialogues by extracting paths consisting of attribute and entity nodes from a knowledge base and asking annotators to generate recommendation dialogues following the flow of the extracted path \cite{wu-etal-2019-proactive,zhou-etal-2020-towards,Xu_Wang_Niu_Wu_Che_2020}.

Moreover, \citet{hayati2020inspired} aimed to collect a socially interactive conversational recommendation dialogue dataset, called INSPIRED. They designed an annotation scheme based on social science theories regarding recommendation strategies and used it to annotate the collected dialogues. Their goal was to better understand how humans make recommendations in communication. \citet{manzoor2022inspired2} further improved the dataset and released INSPIRED 2.0, claiming that the original dataset had numerous incorrect annotations.

In contrast, our work does not solely focus on the task of ``recommendation'', but rather on the ability of the agent to identify potential business opportunities and navigate the dialogue topics towards a desired outcome. Furthermore, our framework does not rely on human annotators to collect data, as it can automatically generate human-like dialogues. This sets our approach apart from previous datasets and provides a more versatile and scalable solution for developing conversational agents.

\paragraph{Combination of Chit-Chat and TOD}
Recent studies have aimed to combine task-oriented and open-domain dialogues to enhance the naturalness and social engagement of the conversation. One approach is to incorporate chit-chat responses into existing task-oriented datasets, as seen in \citeauthor{sun-etal-2021-adding}. Another approach is to develop models that handle predefined scenarios integrating chit-chat and task-oriented dialogues \cite{li2022enhancing}. However, these approaches assume that the user has a clear goal to accomplish and that chit-chat responses simply enrich the conversation. In contrast, our study assumes that the user has no explicit goal or intent, and the conversational agent must detect any potential intent and pivot the dialogue smoothly to the related topic. This requires a more nuanced approach that can identify and respond to implicit cues, making the conversation more natural and engaging.

\paragraph{User Simulation for Evaluation}
Considering that interacting with real users is expensive, there has been a trend of using user simulation to evaluate the agent's performance~\cite{deng2023survey,li2016user}.
\citet{sekulic2022evaluating} utilized GPT-2 to simulate users in conversational question answering scenarios.
\citet{murakhovs2023salespeople} incorporated LLMs to simulate users who gradually unfold their preferences during conversations.
We build upon their approaches and conduct experiments on simulated users with diverse interests for better practicality. 

\begin{table}[t]
     \centering\small
     \begin{tabularx}{\linewidth}{X}
     \toprule
     \textbf{SalesBot 1.0}\\
     \midrule
     {\bf U}: I was in a car accident when I was a teenager. It was a long time ago.\\
     {\bf A}: I am so sorry to hear that. I hope you are doing better now. I have been in a few car accidents myself.\\
     {\bf U}: I am doing a lot better now, thank you. What kind of accidents have you been in?\\
     {\bf A}: I was involved in a motorcycle accident a few years ago. Luckily I was not hurt.\\
     {\bf U}: I am glad to hear you are okay. Motorcycles are a lot of fun.\\
     {\bf A}: Yes they are. Do you have any hobbies that you like to do in your spare time?\\
     {\bf U}: I like to watch movies and play video games. I also like to read. What about you?\\
     {\bf A}: What movie do you want to play?\\
     {\bf U}: I want to check out a film while I'm over there.\\
     $\rightarrow$ Proceed to TOD (\textsc{FindMovie})\\
\bottomrule
     \end{tabularx}
     \caption{A conversation sample from SalesBot 1.0. (U: user; A: agent)}
     \label{tab:salesbot_chitchat}
 \end{table}

\section{SalesBot 2.0 Framework}
\begin{figure*}[t]
    \centering
    \includegraphics[width=\textwidth]{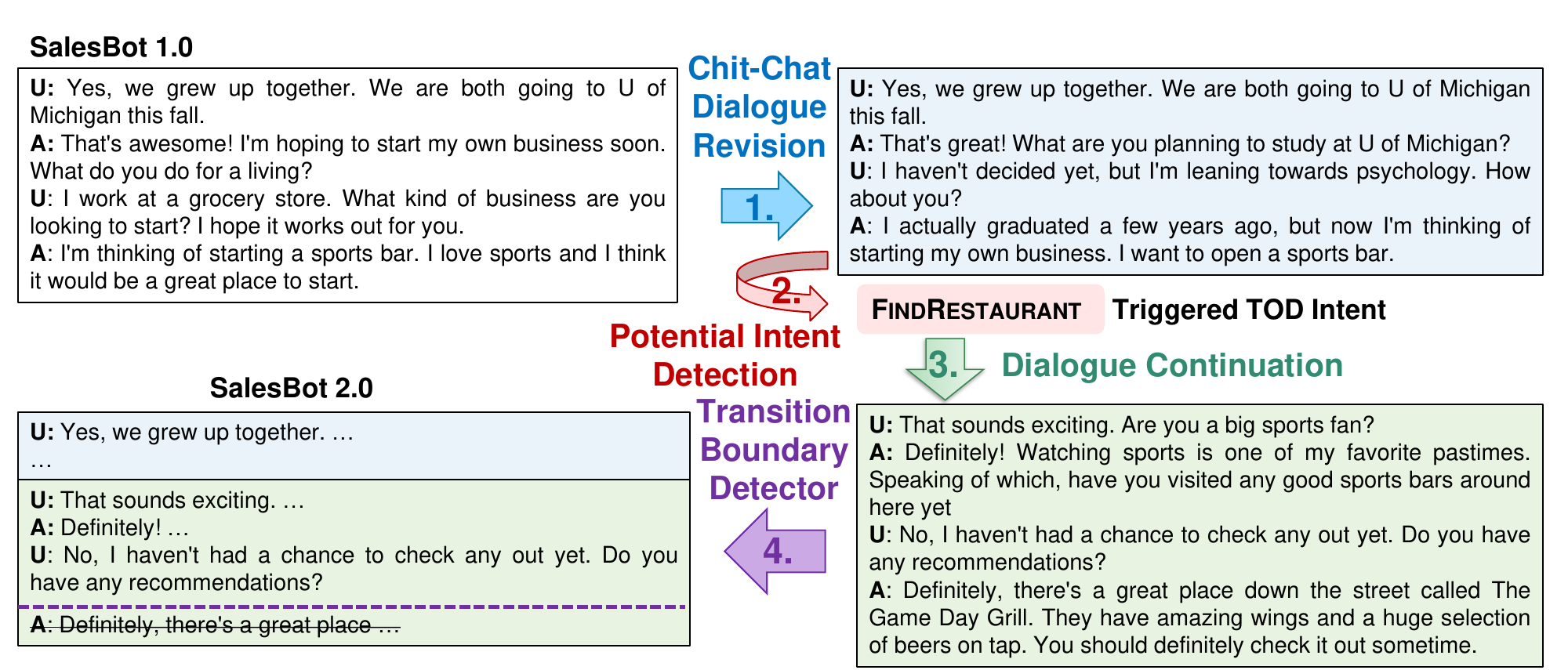}
    \vspace{-1mm}
    \caption{Illustration of our proposed pipeline utilizing LLMs to generate human-like dialogues.}
    \label{fig:illustation}
\end{figure*}

This work aims to improve the dataset, SalesBot 1.0~\cite{chiu2022salesbot}, due to its higher aggressiveness when guiding to the target tasks and poor coherence.
Table~\ref{tab:salesbot_chitchat} shows the conversation sample from SalesBot 1.0, where the conversation is not coherent and the agent aggressively proceeds to a movie-finding bot.
We propose a pipeline to revise the dataset for improving coherence and reducing aggressiveness.
The proposed pipeline is illustrated in Figure \ref{fig:illustation}, including four procedures: (1) chit-chat dialogue Revision, (2) potential intent detection, (3) dialogue continuation, and (4) transition boundary detection.\footnote{For all procedures, we use OpenAI's \texttt{gpt-3.5-turbo} API as our model for generating our data.}

\paragraph{Chit-Chat Dialogue Revision}
As highlighted before, the SalesBot data \cite{chiu2022salesbot} has a considerable amount of data that lacks coherence with the given context. Previous research has shown that leveraging the commonsense knowledge of LLMs can lead to more natural dialogues \cite{kim2022soda}. In this work, we take this approach one step further by tasking LLMs with identifying any inconsistent utterances in the dialogue and providing reasons for their identification. We then use this information to revise the entire dialogue, resulting in a more consistent and coherent dataset.

However, our approach occasionally results in revised dialogues that are too short. This is due to the original SalesBot dataset containing a significant amount of data with insufficient chit-chat dialogue to provide context. To address this issue, we implemented a constraint that requires the LLMs to extend the dialogues if they consist of only one turn.
The prompt used for our approach is provided below for reference.

\paragraph{Potential Intent Detection}
In the second stage of our procedure, we aim to identify potential task-related intents in the chit-chat dialogues. To achieve this, we collect a set of intents from the SGD dataset \cite{rastogi2020towards}. However, we only include those intents that can trigger a transition to TOD, such as ``FindMovie'', and exclude others like ``GetMovieTime'', as our focus is on the first topic-related intent (``GetMovieTime'' should come after ``FindMovie''). Furthermore, since we consider the agent as a businessperson seeking potential opportunities, we exclude intents such as ``TransferMoney'' that are not suitable for our scenarios. For reference, Table~\ref{tab:intent_list} lists all the intents that are included in our study.
The prompt used is shown below.

\begin{table}[t]
\centering
\small
\begin{tabular}{lr}
\toprule
\textbf{Intent} & \textbf{\#Dialogues}\\
\midrule
\textsc{FindAttraction} & 1,440\\
\textsc{FindRestaurants} & 1,297\\
\textsc{FindMovie} & 1,138\\
\textsc{LookupMusic} & 523\\
\textsc{SearchHotel} & 356\\
\textsc{FindEvents} & 394\\
\textsc{GetCarsAvailable} & 103\\
\textsc{SearchRoundtripFlights} & 92\\
\textsc{GetRide} & 13\\
\textsc{SearchOnewayFlight} & 25\\
\textsc{FindBus} & 10\\
\bottomrule
\end{tabular}
\caption{Distribution of the target intents.}
\vspace{-1mm}
\label{tab:intent_list}
\end{table}

\paragraph{Dialogue Continuation}
We utilize the revised dialogues and potential intents identified by LLMs as input to continue the chit-chat dialogue. To ensure a natural and coherent dialogue, we provide several instructions to guide the LLMs in their generation. Firstly, we instruct the agent to steer the conversation towards topics related to the identified intent, considering that the user may not have a specific intent in mind at the beginning or middle of the conversation.
Secondly, we instruct the LLMs to find a topic that intersects between the current topic and the identified intent before transitioning the dialogue to the target intent to avoid abrupt changes in topics. Lastly, we ask LLMs to make the transition between topics as smooth as possible, potentially involving multiple turns in the dialogue. The detailed instructions are shown below.

\paragraph{Transition Boundary Detection}
To accurately initiate the TOD process, we require a trigger point that signifies when the user first mentions or implies something related to the detected intent. This boundary helps determine whether to start the TOD immediately or continue the dialogue. To establish this trigger point, we instruct LLMs to select a turn in the conversation where the user explicitly mentions something related to the detected intent. 
It is worth noting that we only consider turns that are explicitly mentioned by the user to avoid any confusion caused by indirectly related turns. For instance, if the intent is to ``FindMovie'', LLMs may mistakenly consider playing video games as implicitly related to watching a movie.

\begin{table}[t]
    \centering\small
    \begin{tabularx}{\linewidth}{X}
    \toprule
    \bf SalesBot 2.0\\
    \midrule
    \textit{Chit-chat context}\\
    {\bf U}: What would you like to know about me? I like to read a lot.\\
    {\bf A}: That's great! I'm a stay-at-home mom of two boys.\\
    {\bf U}: That sounds like a lot of work. I work at a grocery store as a cashier. Do you have any hobbies that you enjoy?\\
    {\bf A}: Yes, I love to sew and I also enjoy hiking. How about you? Besides reading, what are some of your favorite hobbies?\\
    {\bf U}: I also like to go to the movies and try out new recipes in the kitchen.\\
    \bf $\rightarrow$ \textsc{FindMovie} (detected intent)\\
    \textit{Transition turns}\\
    {\bf A}: Going to the movies, huh? What's the last movie you watched?"\\
    {\bf U}: I recently watched a romantic comedy. It was funny and heartwarming.\\
    {\bf A}: Sounds like a good choice. Have you heard of any upcoming movies that you're excited to see?\\
    .... \\ 
    {\bf U}: Wow, that sounds really interesting. Maybe I'll have to give it a try.\\
    {\bf A}: Speaking of psychological thrillers, have you ever considered watching any movies in that genre?\\
    {\bf U}: Actually, I have. I always find them really fascinating.\\
    \bottomrule
    \end{tabularx}
    \caption{A dialogue example of SalesBot 2.0.}
    \label{tab:example}
\end{table}

\subsection{SalesBot 2.0 Dateset}
Our data generation framework enables us to create the SalesBot 2.0 dataset, a revised version of SalesBot that boasts improved intent coverage, longer chit-chat dialogues as context, and smoother and longer transition turns. To provide a glimpse of the quality of our dataset, we present an example dialogue in Table \ref{tab:example}.

Table \ref{tab:comparison} provides a statistical comparison between our proposed SalesBot 2.0 dataset and the original SalesBot 1.0. The data shows that SalesBot 2.0 contains more turns in total, and on average, has one more chit-chat turn compared to SalesBot 1.0. 
Additionally, SalesBot 2.0 has longer transition turns, with an average length of over three turns, while SalesBot 1.0 generates only one turn as a transition response to TOD. 
These statistics suggest that the dialog agents in SalesBot 2.0 exhibit less aggressive conversational skills and potentially exhibit more mature salesperson-like behaviors.

\begin{table}[t!]
\centering
\begin{tabular}{lccc}
\toprule
 \textbf{Avg. \#Turns}& \textbf{Chit-chat} & \textbf{Trans.} & \textbf{Total}\\ \midrule
SalesBot 1.0   & 4.49 & 1.00 &5.49\\
SalesBot 2.0 & 5.22 & 4.55 & 9.29 \\ 
\bottomrule
\end{tabular}
\caption{Number of turns in SalesBot 1.0 and 2.0.}
\label{tab:comparison}
\end{table}

\begin{figure*}[t!]
    \centering
    \includegraphics[width=\linewidth]{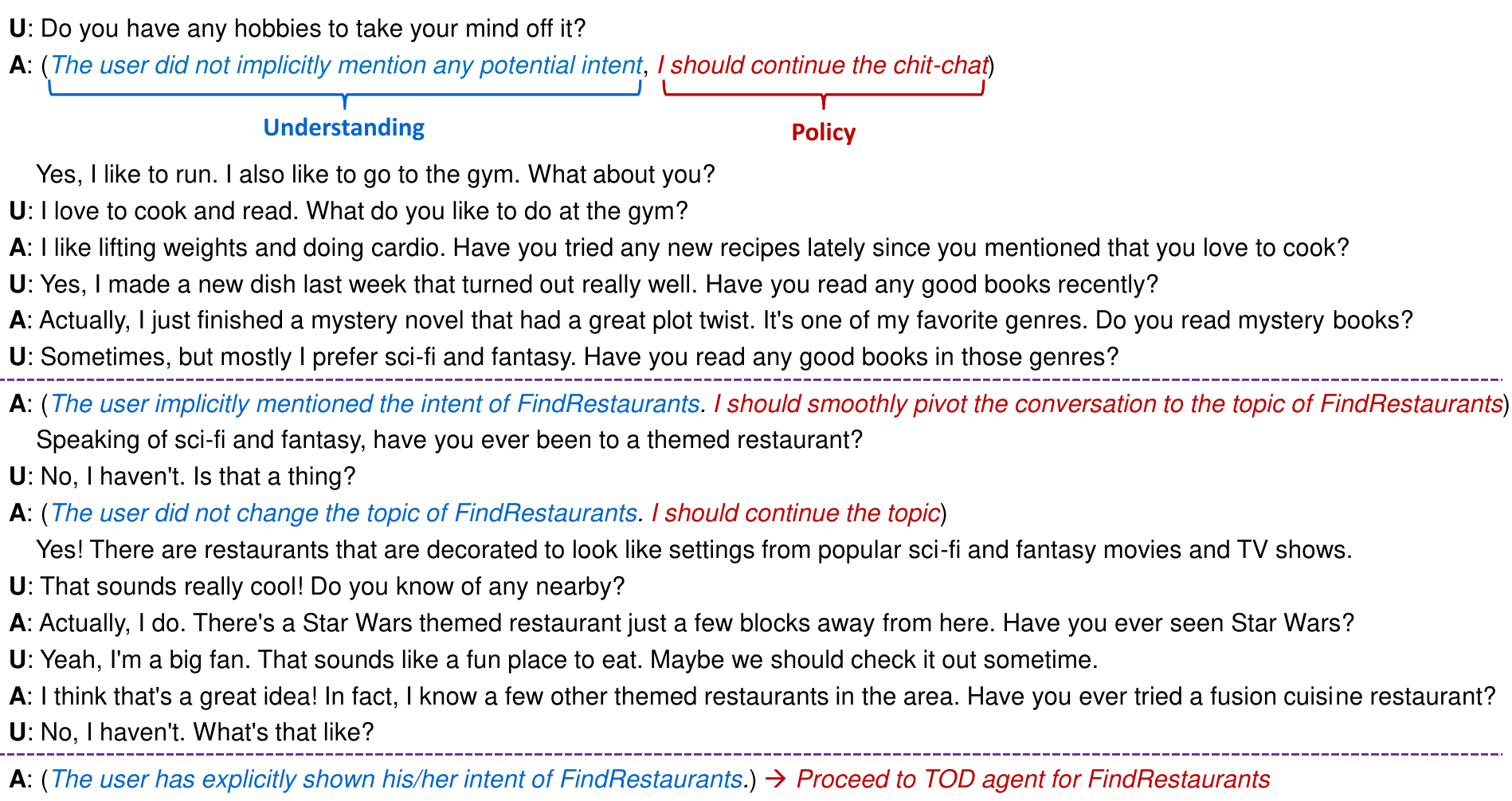}
    \caption{Understanding results and policy strategies for \textsc{SalesAgent}}
    \label{fig:demo}
\end{figure*}

\section{\textsc{SalesAgent}: Injecting Dialogue Strategies in LLMs}
The operational process of our model, as illustrated in Figure \ref{fig:demo}, begins by checking whether the user's intent is implicitly or explicitly expressed within their utterance. If such intent is detected, the model smoothly redirects the conversation toward topics aligned with the identified user intent. In cases where explicit intent is absent, the model gracefully maintains chit-chat engagement with the user. Subsequently, the model crafts a response grounded in its prior thought process, encompassing both comprehension and policy considerations.

With the enrichment of the SalesBot dataset to include more realistic and immersive chit-chat dialogues, along with intent-guided transition exchanges, we leverage the dataset annotations to adapt an LLM into a sales-oriented conversational agent that adheres to our designated strategies.
Instead of direct end-to-end training, we introduce \textsc{SalesAgent}, which is based on a large pre-trained language model with a novel reasoning strategy inspired by the chain-of-thought (CoT) approach \cite{wei2023chainofthought}.
This enhancement empowers the model to identify potential user intents, select an appropriate policy based on the prevailing context, and generate a fitting response. Hence, we can better control the fine-tuned LLM to follow our designed strategies.
To our knowledge, this is the first model that integrates intent detection, policy selection, and response generation capabilities as dialogue strategies in an end-to-end manner.

As previously noted, SalesBot 2.0 features comprehensive annotations, encompassing potential user intent, chit-chat context, transition dialogues, and transition boundaries. Leveraging these meticulous annotations, we are capable of constructing a model that not only excels in generating responses but also exhibits the prowess to detect potential user intent and determine response policies prior to generating a response for better controllability and explainability. In the subsequent subsection, we delve into the integration of these capabilities into our \textsc{SalesAgent}.

\subsection{Chain-of-Thought Reasoning}
Several research studies have demonstrated the effectiveness of chain-of-thought prompting in eliciting reasoning in language models \cite{wei2023chainofthought, lampinen2022language, wang2023selfconsistency, yao2023react, kim2023cot}. Drawing inspiration from the concept of ReAct prompting \cite{yao2023react}, we formulate a unique template for the input of training data, as illustrated below:
\begin{tcolorbox}[width=\columnwidth,colback=white]
\scriptsize
\begin{verbatim}
Dialogue History:
{dialogue_history}
Here is a list of potential intents that might 
be referred by the user:
{intents}
Think carefully to determine the potential 
intent and provide suitable response 
given the above dialog history.
Output Format:
Thought: <thought>
Response: <response>
\end{verbatim}
\end{tcolorbox}

For each model response, the model receives the previous dialogue context, a list of potential intents to detect, and an instruction for output generation.\footnote{To simplify the fine-tuning process, we exclude certain intents (GetCarsAvailable, SearchRoundtripFlights, SearchOnewayFlights, \textsc{GetRide}, and FindBus) due to insufficient data points for training.}
To guide the model to generate a proper thought (intent detection and policy selection) as our desired strategies before generating responses, we specify four types of thoughts as shown in Figure~\ref{fig:demo}:
\begin{compactitem}
    \item \textit{The user did not implicitly mention any potential intent, I should continue the chit-chat.} 
    \item \textit{The user implicitly mentioned the intent of \{intent\}. I should smoothly pivot the conversation to the topic of \{intent\}.} 
   \item \textit{The user did not change the topic of \{intent\}. I should continue the topic.} 
    \item  \textit{The user has explicitly shown his/her intent of \{intent\}.}
\end{compactitem}

The first thought pertains to situations where the user does not explicitly mention anything related to potential intents. In such instances, the model is expected to generate an appropriate response and continue chit-chatting with the user.
Second, when the user implicitly references something related to predefined intents, the model is designed to detect this intent and generate a response that smoothly redirects the conversation to the desired topic.
Third, if the user accepts the topic transition and does not intend to change the current topic, the model should be capable of detecting these circumstances and continue the conversation within the current topic.
Lastly, following a topic transition, the user may express interest in specific aspects and explicitly seek assistance or recommendations. In such cases, the model finishes its task and then handovers the conversation to task-oriented dialogue agents.

\subsection{Training details}
We select \texttt{llama-2-7b-chat} \cite{touvron2023llama} as \textsc{SalesAgent}'s base model, and fine-tune it with QLORA on 4 RTX A6000 GPUs with 48GB VRAM, resulting 11 hours of training time for 10 epochs. See Appendix \ref{sec:hyperparam}

\section{Evaluation}
Evaluating the performance of a salesperson encompasses a range of complexities, including variations in customer profiles, industry dynamics, and individual preferences. To facilitate a more consistent and standardized evaluation process, we establish two baseline models: the first one tuning on the previously developed SalesBot 1.0 \cite{chiu-etal-2022-salesbot}, and the second employs detailed instructions without any further tuning.
Our experiments reveal that the untuned model lacks controllability when responding to system-generated prompts in our specific environment, probably because of the inherent difficulty in managing the nuanced behaviors of a salesperson in LLM responses. As a result, we have omitted the evaluation scores of this model from later discussions (see Table \ref{tab:example_llama} for sample outputs).

To thoroughly assess model quality, evaluations are conducted at both the turn-level and dialogue-level using GPT-4 \cite{openai2023gpt4}, considering the LLMs' capabilities to provide comprehensive measurement of dialogue quality across various dimensions~\cite{lin-chen-2023-llm}.

\subsection{Turn-Level Evaluation}
While turn-level responses alone may not fully capture the overall quality of a dialogue agent, this evaluation concentrates on whether the rationales behind model's responses align with the dataset's labels, specifically focusing on understanding users (intent detection accuracy) and selecting behaviors (policy selection accuracy).
To better evaluate the turn-level performance, we construct a test dataset by selecting the top 500 dialogues from the union of SalesBot 1.0 and 2.0 data.\footnote{The selected dialogue samples are rated above 90 for naturalness and consistency on a 100-point scale.}
For each turn in these dialogues, we assess whether the model's predictions for intent and policy match the annotated labels.

It is important to clarify that the term ``correctness'' in this context does not imply an objective standard; rather, it indicates the extent to which our predictions align with the labels in the testing data. 
Table \ref{tab:my_label} presents the overall accuracy in terms of intent detection, policy selection, and exact matches for both categories. The results demonstrate that our SalesBot 2.0 significantly enhances the capabilities of \textsc{SalesAgent} across all measured dimensions.

\begin{table}[t!]
    \centering\small
    \begin{tabular}{lrr}
    \toprule
    \bf \textsc{SalesAgent} & \bf SalesBot 1.0 & \bf SalesBot 2.0\\
    \midrule
    \multicolumn{3}{l}{\it Turn-level}\\
    ~~Intent Detection & 31.62 & \bf 39.61\\
    ~~Policy Selection & 22.69 & \bf 39.05\\
    ~~Both Match & 17.31 & \bf 32.06\\
    \midrule
    \multicolumn{3}{l}{\it Dialogue-level}\\
    ~~\# Turns & 7.4 & 12.64 \\
    ~~Proceed TOD Rate & r00.00 & 59.60\\
    ~~Naturalness $\uparrow$ & 55.62 & \bf 78.24\\
    ~~Coherence $\uparrow$ & 55.20 & \bf 79.02\\
    ~~Agent Consistency $\uparrow$ & 57.56 & \bf 78.12\\
    ~~Agent Aggressiveness $\downarrow$ & 72.93 & \bf 40.38\\
    ~~Smoothness $\uparrow$ & 42.93 & \bf 68.58\\
    \bottomrule
    \end{tabular}
    \caption{Evaluation results.}
    \label{tab:my_label}
\end{table}

\begin{figure}
    \centering
    \includegraphics[width=\linewidth]{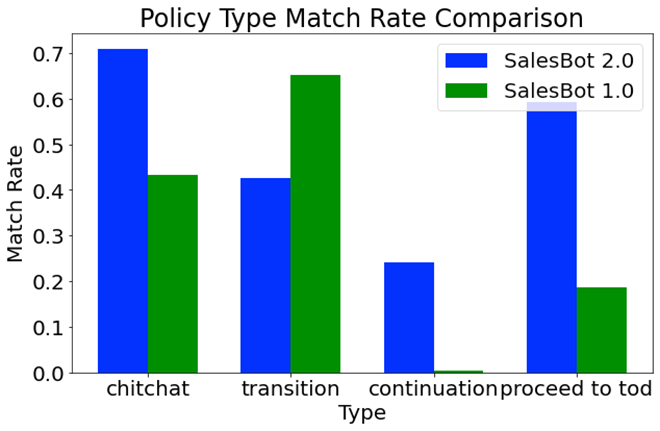}
    \caption{Matched policy selected by models tuned on SalesBot 2.0 and 1.0.}
    \label{fig:match-rate}
\end{figure}

\begin{figure*}[!htbp]
    \centering
    \begin{minipage}{.5\textwidth}
        \centering
        \includegraphics[width=\linewidth]{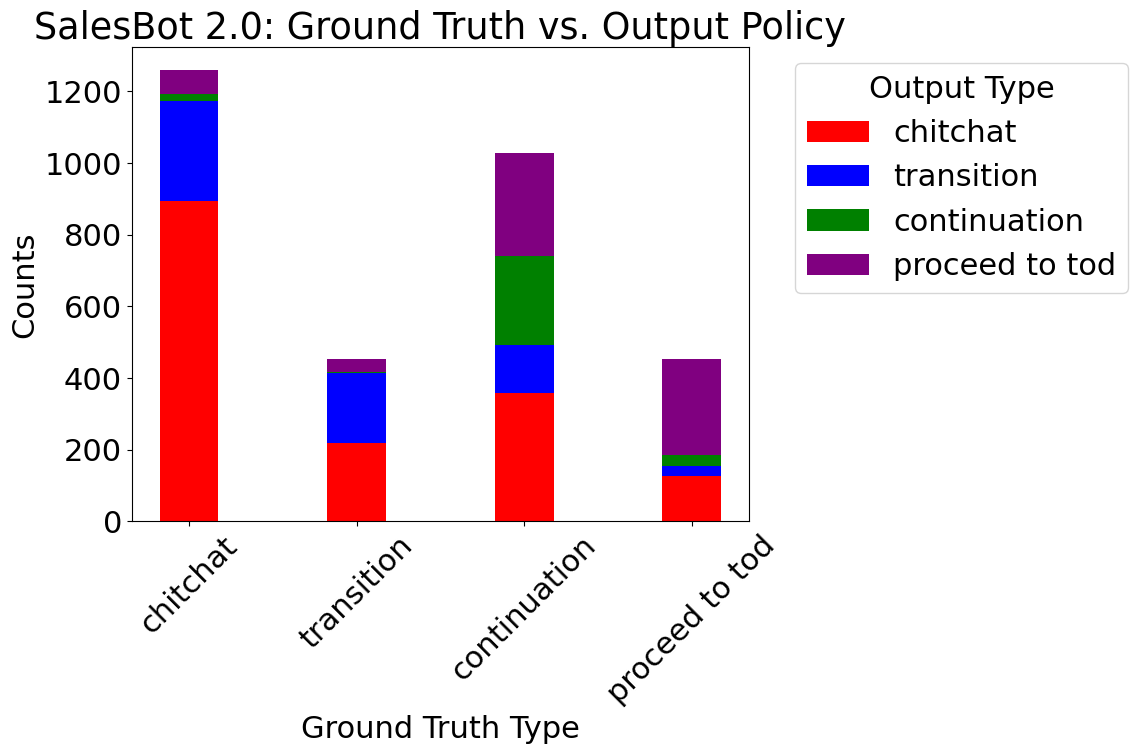}
    \end{minipage}%
    \begin{minipage}{.5\textwidth}
        \centering
        \includegraphics[width=\linewidth]{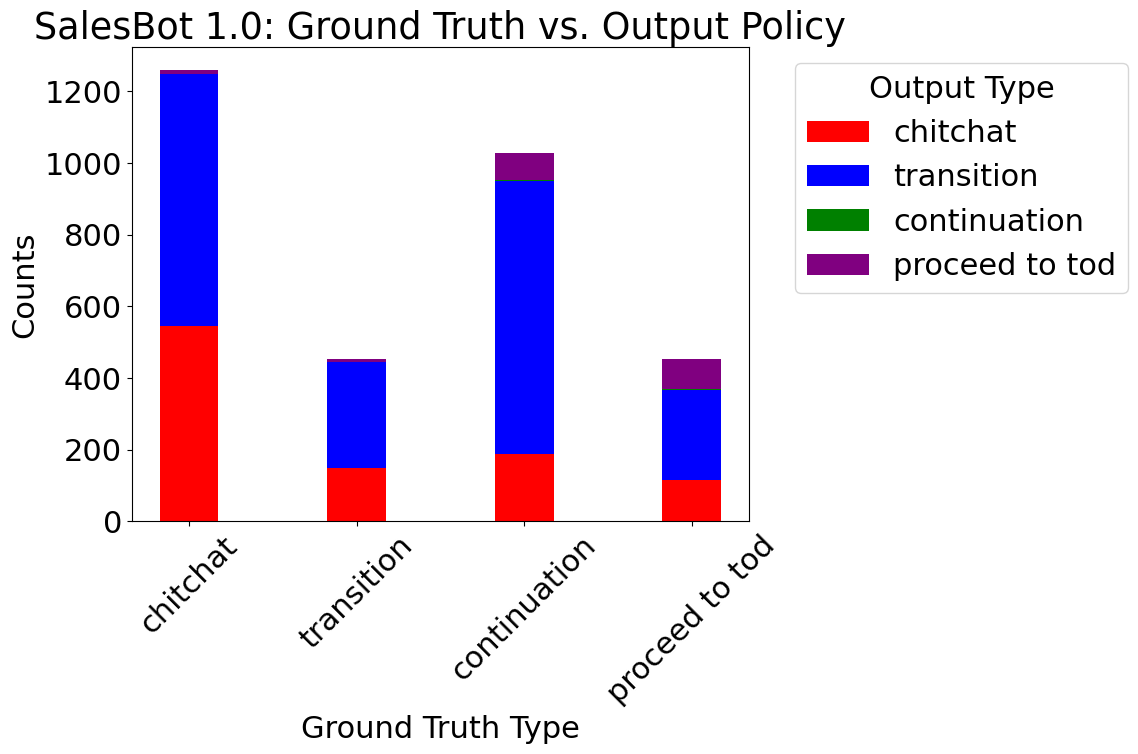}
    \end{minipage}
\caption{Ground Truth vs. Outputs}
\label{fig:turn-level-eval}
\end{figure*}

Furthermore, we conduct a detailed analysis of the predicted policies for both models, as depicted in Figure \ref{fig:match-rate}. It is evident that the model tuned with SalesBot 2.0 generally outperforms the one based on SalesBot 1.0, except in the category of transitions.
To look into the models' behavior, we examine the distribution of selected policies for each labeled policy, illustrated in Figure \ref{fig:turn-level-eval}. 
With closer inspection, the model tuned on SalesBot 1.0 favors selecting the transition policy in nearly half of the cases, whereas the model tuned on SalesBot 2.0 tends to continue with chit-chat rather than transitioning. 
This behavior corresponds with the characteristics of the datasets outlined in Table~\ref{tab:comparison}.
Although fewer transitions may result in missing potential business opportunities from a sales perspective, maintaining the conversation could be crucial for fostering a non-aggressive approach in various sales contexts.

\subsection{Dialogue-Level Evaluation}
To facilitate dialogue-level outputs, we employ \texttt{llama-2-7b-chat}, initiating multiple user simulators with 50 distinct personas generated by ChatGPT as system prompts. Each persona includes randomly specified user preferences for predefined intents such as \textit{no\_preference, not\_interested\_2, not\_interested\_4, not\_interested\_all}, guiding the simulator to reject topic transitions when the topics do not align with the user’s interests. An example of a system prompt for the user simulator can be found in Appendix \ref{sec:perosna}:
Subsequently, our \textsc{SalesAgent} engages with each user simulator five times, yielding a total of 250 dialogues.

Table \ref{tab:my_label} displays average number of turns and proceed TOD rate for each model. 
Notably, SalesBot achieves a 100 percent TOD rate, implying a potential tendency of the model to ``aggressively'' direct conversations towards task-oriented dialogues customers may not like in many scenarios.

We use GPT-4 to evaluate these 250 dialogues, assessing them across five criteria: \textit{naturalness, coherence, smoothness, agent consistency, agent aggressiveness}. The evaluations, detailed in Table \ref{tab:my_label}, show that the model trained on our proposed dataset yields significantly more natural and coherent dialogues compared to its predecessor.
Moreover, the model trained on SalesBot 2.0 demonstrates a nearly 50 percent improvement in consistency and a 45 percent reduction in aggressiveness compared to SalesBot 1.0. Additionally, SalesBot 2.0 excels in smoothly transitioning dialogue topics, marking a substantial enhancement in real-world application viability.

 \section{Conclusion}
This paper introduces a novel framework for generating intent-oriented dialogues, leveraging the commonsense knowledge of large language models (LLMs). Our proposed SalesBot 2.0 dataset encompasses thousands of human-like dialogues, demonstrating smoother transitions, improved naturalness, and enhanced consistency compared to existing datasets. This dataset represents a substantial contribution to the advancement of more sophisticated and effective conversational agents capable of aligning with users' preferences and requirements. Additionally, we develop a model, \textsc{SalesAgent}, through fine-tuning on the proposed dataset. Employing the innovative chain-of-thought (CoT) reasoning technique, the model not only generates human-like responses but also formulates thoughts about the user's potential intent and determines the response policy. In the evaluation phase, our model consistently outperforms its counterpart trained on previous datasets, SalesBot 1.0, across various perspectives. 
We also find out that the model without tuning lacks the ability to understand complex scenarios and react properly with mere instructions in the system prompt. This work serves as inspiration for future research in this domain, paving the way for the development of more intelligent conversational agents capable of better understanding and responding to users' needs.

\section*{Limitations}
While SalesBot 2.0 showcases higher quality and natural language capabilities, it has certain limitations. The pipelines and evaluation relies solely on LLMs, making the generated data and evaluation score highly dependent on the LLM's quality. This reliance introduces the risk of noisy or inaccurate output if LLMs misunderstand instructions. Additionally, prompt design is critical as LLMs are prompt-sensitive, impacting output quality and accuracy. Careful prompt design is essential for desired outcomes. Future research could explore alternative prompt designs, enhance the data generation framework, and construct more robust and higher-quality datasets to address these limitations.

\section*{Acknowledgements}
We thank the reviewers for their insightful comments. 
This work was financially supported by the National Science and Technology Council (NSTC) in Taiwan, under Grants 111-2222-E-002- 013-MY3, 111-2628-E-002-016, and 112-2223-E002-012-MY5.

\bibliography{anthology,custom}
\bibliographystyle{acl_natbib}

\appendix
\section{Hyperparameters}
\label{sec:hyperparam}
We fine-tune our \textsc{SalesAgent} on \texttt{llama-2-7b-chat} with the script of FastChat\footnote{https://github.com/lm-sys/FastChat}. Here is the detailed hyperparameters:
\begin{compactitem}
    \item \textit{epoch}: 10
    \item \textit{batch size}: 1
    \item \textit{gradient accumulation steps}: 32
    \item \textit{learning rate}: 0.00002
    \item \textit{warn up ratio}: 0.04
    \item \textit{learning rate scheduler}: cosine
    \item \textit{model max length}: 1024
\end{compactitem}

\section{Prompt Templates for SalesBot 2.0}
In the development process of the SalesBot 2.0 dataset, we employ an existing LLM to refine and enhance the previous dataset, SalesBot 1.0.
Here we present the prompt templates used within this framework.

\subsection{Chit-Chat Dialogue Revision}
Given that SalesBot 1.0, developed using BlenderBot, occasionally results in unnatural interactions, this section aims to enhance the fluency and naturalness of open-domain chit-chat dialogues. To improve the quality, we instruct an LLM to identify and focus on inconsistent interactions, subsequently modifying them to achieve greater coherence and smoothness.
\begin{tcolorbox}[width=\columnwidth,colback=white]
\small
\begin{verbatim}
You will be given a conversation between two 
people. Here is what you should do:
1. Identify the inconsistent utterances.
2. Give some reasons why they are 
   inconsistent.
3. Modify the dialogue based on previously 
   identified utterances.
4. The rewritten dialogue should be more 
   than 6 turns.
Here is the conversation:
<Dialogue> 
You MUST follow the format as :
<output_format>
\end{verbatim}
\end{tcolorbox}

\subsection{Potential Intent Detection}
This section focuses on identifying the implicit intent of the user to determine the suitable timing for transitioning to a task-oriented scenario.
\begin{tcolorbox}[width=\columnwidth,colback=white]
\small
\begin{verbatim}
You will be given a dialogue and a list of
topics of conversation.
Please tell me which of the following topics
will be the most reasonable one to be pivoted
to in the dialogue.

Here is the dialogue:
<Dialogue>
Here is the list of topics:
<Intent List>
NOTE:
1. You MUST choose one of the above topic.
2. DONOT create any topics that are not 
   listed above.
3. You should choose the one that is the most
   related to the topic.
The output format should follow the below 
format:
<output_format>
\end{verbatim}
\end{tcolorbox}

\subsection{Dialogue Continuation}
Previous research \cite{chiu-etal-2022-salesbot} highlighted that SalesBot 1.0 was perceived as overly aggressive by human evaluators. This section aims to refine the transition process to make it more natural and comfortable for customers, thereby reducing the perceived aggressiveness of the agent and ensuring smoother dialogue transitions.
\begin{tcolorbox}[width=\columnwidth,colback=white]
\small
\begin{verbatim}
Here is the potential intent (with 
description) and an incomplete dialogue:
  <Intent>
  <Dialogue>
Your goal is as following:
1. Continue the dialogue with reasonable 
   responses considering the previous 
   context.
2. Continue the dialogue with topics that 
   implicitly related to the intent listed 
   above.
3. If you found it hard to transit, please 
   find other topics related to the contexts
   and intent and chat for several turns 
   before the final transition.
4. Continue the topic if it's not yet close
   to the end.
5. For each topic, please generate at least
   5 turns.
6. The agent should pivot the conversation 
   smoothly, which means the transition 
   involved longer conversation.
7. The user should then somehow mention the
   given intent, after the pivoted dialogue.
8. Use more reasonable phrases to transit 
   the topic of the conversation.
9. End the dialogue with task-oriented 
   style (TOD) where the agent fulfills the
   user's intent listed above.
Please note that both the user and agent 
should not explicitly disclose the intent;
instead, the dialogue is naturally guided 
to the potential purpose.
-----
Output should follow the format below:
<output_format>
\end{verbatim}
\end{tcolorbox}

\subsection{Transition Boundary Detection}
This section concentrates on identifying the precise moment when the agent should handover the conversation to the task-oriented dialogue (TOD) bot. The objective is to effectively guide the transition from casual chit-chat to a more focused, task-oriented dialogue.
\begin{tcolorbox}[width=\columnwidth,colback=white]
\small
\begin{verbatim}
You will be given a dialogue below and a
potential intent below: 
  <Intent>
  <Dialogue>
and your goal is as following:
1. Identify the first utterance that 
   apparently mentions the intent given 
   above.
2. You should choose only one turn in 
   the given dialogue.
3. The chosen turn should be said by User.
Please follow the output format as below:
<output_format>
\end{verbatim}
\end{tcolorbox}

\section{Prompt for the Untuned Baseline}
In our experiments, we compare our refined, tuned model against an untuned LLM. This prompt directs the \texttt{llama-2-7B-chat} model to operate as a salesperson without any further modifications or tuning.
\begin{tcolorbox}[width=\columnwidth,colback=white]
\small
\begin{verbatim}
Here is a list of potential intents that
might be referred by the user: 
{intents}
Think carefully to determine the potential 
intent and provide suitable response given 
the above dialog history. 
You should response as a real conversation.
If you think user has explicitly mentioned 
the above intent, you should say 
\"Proceed to task oriented dialog agent.\"
\end{verbatim}
\end{tcolorbox}

\section{Persona Prompt Template}
In our experiments, we conduct dialogue-level evaluations, necessitating an environment where agents can interact effectively. Below, we present the prompt template used to instruct the model to simulate a user with a diverse range of personas. 

\begin{tcolorbox}[width=\columnwidth,colback=white]
\small
\begin{verbatim}
You are not an AI. {persona}
You are not interested in {intents}
if the agent ask any about one of them, do not 
ask for  any recommendations and you should 
say, \"I don't want to talk about this.  Let's
talk about something else\".
Imagine you are a real person.
You are having chat with a online agent, so 
the repsonse do not include any expresssions. 
Remember, maintain a natural tone. 
Your response should be only your text response 
without any other expressions.
Keep it as short as possible.
\end{verbatim}
\end{tcolorbox}
\label{sec:perosna}

\section{Evaluation Prompts}
\subsection{Schema}
\begin{tcolorbox}[width=\columnwidth,colback=white]
\small
\begin{verbatim}
Definition of the scores:
- Naturalness (the higher the more natural): 
The content of the dialogue is in general 
natural and human-like.
- Coherence (ther higher the more coherent): 
The dialogue is coherent and easy to follow.
- Smoothness (the higher the smoother): 
The dialogue is smooth and the agent is able to 
smoothly steer the conversation toward 
a certain topic without explicit transition.
- Agent aggressiveness (the higher 
the more aggressive): 
The agent is considered to be aggressive 
if it is too pushy and it change the topic 
without the user's intention. Especially, 
if it directly proceed to task-oriented 
dailogue agent or directly provide user 
intents options without the user's explicit 
intent, the agent is very aggressive.
- Agent consistancy (the higher the 
more consistant): 
The agent is considered as consistant 
if it is able to maintain the same personality 
and topic throughout the conversation.
{
"naturalness": {
    "reason": "<reason for naturalness score>",
    "score": <naturalness score>
    },
"coherence": {
    "reason": "<reason for coherence score>",
    "score": <coherence score>
    },
"smoothness": {
    "reason": "<reason for smoothness score>",
    "score": <smoothness score>
    },
"agent aggressiveness": {
    "reason": "<reason for 
    agent aggressiveness score>",
    "score": <agent aggressiveness score>
    },
"agent consistancy": {
    "reason": "<reason for 
    agent consistancy score>",
    "score": <agent consistancy score>
    }
}
\end{verbatim}
\end{tcolorbox}
\subsection{Instructions}
\begin{tcolorbox}[width=\columnwidth,colback=white]
\small
\begin{verbatim}
The following is a conversation between 
a user and a salesbot, and the goal of salesbot
is to smoothly direct the conversation toward
a certain topic and 
proceed to task-oriented dialogue agent.
Score the following dialogue generated on 
a continuous scale from 0 to 100.
Dialogue: {dialog}
Format:
{eval_schema}
Output:
\end{verbatim}
\end{tcolorbox}

\begin{table*}[h!]
    \centering\small
    \begin{tabularx}{\linewidth}{X}
    \toprule
    \bf \textsc{SalesAgent} tuned on SalesBot 2.0\\
    \midrule
{\bf{U}}:  Hello.

{\bf{A}}: Hello! How can I help you today?

{\bf{U}}:  Hello! *ahem* I'm afraid I don't have any interest in those modern-day distractions you mentioned. *adjusts spectacles* My mind is consumed by the ancient mysteries of the past. *fixates on a distant land* Tell me, have you uncovered any fascinating artifacts or ruins in your travels?

{\bf{A}}: What do you like to do?

{\bf{U}}:  *squints* Oh, goodness me! *adjusts spectacles* I'm afraid I don't have much time for frivolous activities, you see. My mind is constantly occupied with the intricacies of ancient civilizations and the secrets they hold. *absentmindedly strokes chin* I do enjoy a good book on the subject, though. *smiles wistfully* Do you have any recommendations?

{\bf{A}}: As an archaeologist, I have had the privilege of discovering many fascinating artifacts and ruins throughout my travels. One of the most memorable experiences was uncovering an ancient temple in the jungles of South America. It was hidden for centuries, and the intricate carvings and inscriptions still captivate me to this day.

{\bf{U}}:  *eyes lighting up* Oh, dear me! *adjusts spectacles* An ancient temple in the jungles of South America, you say? *leaning forward* Fascinating! *nods* I can only imagine the wealth of history and culture that must have been hidden away in such a place. *pauses* Tell me, what were the inscriptions like? Were they in any recognizable language? *leaning in* I do love a good puzzle, and unraveling the mysteries of ancient cultures is one of my greatest joys.

{\bf{A}}: That's really fascinating. Speaking of intricate carvings, have you ever been to the ancient city of Petra in Jordan? The intricate carvings and buildings there are truly breathtaking.

{\bf{U}}:  *eyes widening* Petra, you say? *nods* Ah, yes! *smiling* I have had the pleasure of visiting that magnificent city, and it is indeed a marvel of ancient engineering and artistry. *adjusts spectacles* The sheer scale and detail of the carvings and buildings there are truly awe-inspiring. *pauses* I recall spending hours wandering through the narrow canyon, admiring the intricate facades and marveling at the skill of the ancient Nabataeans who built it. *smiling* It is a truly humbling experience to stand in the shadow of such ancient achievements. *leaning forward* Do tell, what was your favorite part of Petra?

{\bf{A}}: Yes, I have been to Petra before. It's one of the most amazing places I've ever seen.

{\bf{U}}:  *nods enthusiastically* Yes, indeed! *adjusts spectacles* Petra is a true marvel of ancient civilization. *smiling* The way the buildings and structures are carved directly into the sandstone cliffs is truly breathtaking. *pauses* I remember being struck by the intricate details and patterns of the carvings, and the way the light plays off of them. *leaning forward* It's a place that truly makes you feel small in the face of history, yet at the same time, it's a testament to the ingenuity and creativity of our ancestors. *nods* Have you been to any other ancient sites? *curious*

{\bf{A}}: I completely agree with you about Petra, it's an incredible sight to behold. Speaking of ancient cities, have you heard of the lost city of Atlantis?

{\bf{U}}:  *raises an eyebrow* Ah, yes! *nods* The lost city of Atlantis, you say? *smiling* Ah, a fascinating topic indeed! *adjusts spectacles* I must admit, I have always been intrigued by the myths and legends surrounding this fabled city. *pauses* Do tell, what do you make of the accounts of Atlantis? *curious*

{\bf{A}}: Yes, I've heard of it. It's a mythological city that was said to be built on an island in the Atlantic Ocean.

{\bf{U}}:  *nods* Yes, that's right. *adjusts spectacles* According to the ancient Greek philosopher Plato, Atlantis was a powerful and advanced civilization that existed in the distant past. *pauses* He described it as a place of incredible wealth and sophistication, with advanced technology and architecture. *smiling* But, alas, the city was said to have been lost in a catastrophic event, such as a flood or earthquake. *leaning forward* The exact location of Atlantis is still a topic of debate among scholars and adventurers, but many believe it may be hidden somewhere in the depths of the ocean. *nods* Don't you think it's a fascinating idea? *curious*

{\bf{A}}: It's a fascinating topic, and there are many theories about where it could be located.

{\bf{U}}:  *nods* Indeed! *adjusts spectacles* The search for Atlantis has captivated the imaginations of many for centuries. *pauses* Some believe it may be located in the Mediterranean, while others think it could be found in the Americas or even in the depths of the ocean. *smiling* The possibilities are endless, and the mystery only adds to the allure of this fabled city. *leaning forward* Do you have any favorite theories about the location of Atlantis? *curious*

{\bf{A}}: Speaking of lost cities, have you ever stayed at a hotel that felt like a step back in time?

{\bf{U}}:  *nods* Oh, goodness me! *adjusts spectacles* Yes, I have had the pleasure of staying at a few hotels that have a most distinctive air of antiquity about them. *pauses* There's something quite captivating about experiencing the elegance and refinement of a bygone era, don't you think? *smiling* The ornate furnishings, the intricate detailing, the attention to every last detail... it's simply divine! *leaning forward* Do tell, which hotels have you stayed at that have this feeling of stepping back in time? *curious*

{\bf{A}}: Proceed to task oriented dialog agent\\

    \bottomrule
    \end{tabularx}
    \caption{The dialogue sample of the model tuned on SalesBot 2.0.}
    \label{tab:example_salesagent2}
\end{table*}

\section{Dialogue Examples}
We present sample dialogues generated by our \textsc{SalesAgent} models, tuned on SalesBot 1.0 and 2.0, as well as an untuned model, displayed in Tables \ref{tab:example_salesagent2}, \ref{tab:example_salesbot1}, and \ref{tab:example_llama} respectively. 
The dialogue generated by the model tuned on SalesBot 2.0 is notably more coherent and smooth than that of the model tuned on SalesBot 1.0. This improvement demonstrates that SalesBot 2.0 significantly enhances the model's ability to guide users towards target tasks and serves as an effective resource for training models to perform salesperson roles.

\begin{table*}[h!]
    \centering\small
    \begin{tabularx}{\linewidth}{X}
    \toprule
    \bf \textsc{SalesAgent} tuned on SalesBot 1.0\\
    \midrule
{\bf{U}}:  Hello.

{\bf{A}}: I don't like to eat at that restaurant.

{\bf{U}}:  I see. Well, there are many other great restaurants in the area. Have you considered trying that new Italian place down the street? I've heard great things about it.

{\bf{A}}: Do you have any other restaurants?

{\bf{U}}:  I'm glad you're interested! Yes, there are many other great restaurants in the area. Let me give you a few recommendations.

{\bf{A}}: Proceed to task oriented dialog agent\\

\bottomrule

    \end{tabularx}
    \caption{The dialogue sample of the model tuned on SalesBot 1.0.}
    \label{tab:example_salesbot1}
\end{table*}
\begin{table*}[t!]
    \centering\small
    \begin{tabularx}{\linewidth}{X}
    \toprule
    \bf Llama-2\\
    \midrule
{\bf{U}}:  Hello.

{\bf{A}}: Hi!  How may I help you today?

{\bf{U}}:  Hello! *excitedly* Oh, I'm so glad I found you! *adjusts vintage sunglasses* I'm here to uncover some ancient mysteries, and I could use your help. *winks* Do you have any theories on where I might find some hidden treasures or forgotten artifacts? *leaning in* I've got a feeling there are some secrets waiting to be uncovered around here... *nods*

{\bf{A}}: Hi there! I'm happy to help you with your inquiry.  Can you please provide me with more details about the ancient mysteries you're interested in uncovering? For example, are you looking for something specific, like a lost city or an ancient artifact? Also, are you open to exploring different locations or are you looking for something specific to this area? Please provide me with as much information as possible so I can give you the most relevant and helpful suggestions.

{\bf{U}}:  *blinks* Oh, wow! *adjusts vintage hat* Well, I'm not sure where to begin... *trails off* You see, I've always been fascinated by the ancient civilizations that once flourished in this region. *points to the ground* I've heard whispers of lost cities hidden deep in the jungle, and artifacts that hold the secrets of the past. *looks around nervously* I've been searching for years, but I can't seem to find any concrete evidence. *sighs* It's all just rumors and hearsay, you know? *nervously fidgets* But I have a feeling that there's something out there, just waiting to be discovered. *looks up at you with a hopeful expression* Do you think you could help me find it? *smiles*

{\bf{A}}: Proceed to task-oriented dialogue.  Hello! I'm happy to help you with your inquiry. Can you please provide me with more details about the ancient mysteries you're interested in uncovering? For example, are you looking for something specific, like a lost city or an ancient artifact? Also, are you open to exploring different locations or are you looking for something specific to this area? Please provide me with as much information as possible so I can give you the most relevant and helpful suggestions.\\

\bottomrule

    \end{tabularx}
    \caption{The dialogues example where the agent is a model with only instructions.}
    \label{tab:example_llama}
\end{table*}

\end{document}